\documentclass{article}
\usepackage{spconf,amsmath,graphicx}
\usepackage{graphicx}
\usepackage{epstopdf}
\usepackage[misc]{ifsym}
\usepackage{enumitem}
\usepackage{amsmath}
\usepackage{amssymb}
\usepackage{url}
\usepackage{multirow}
\usepackage{subfig}

\title{Fast and accurate reconstruction of HARDI using a 1D encoder-decoder convolutional network}
\name{Shi Yin, Zhengqiang Zhang, Qinmu Peng, Xinge You}
\address{School of Electronic Information and Communications,\\Huazhong University of Science and Technology, Wuhan, China}
\begin{document}
%
\maketitle
\begin{abstract}
High angular resolution diffusion imaging (HARDI) demands a lager amount of data
measurements compared to diffusion tensor imaging, restricting its use in practice.
In this work, we explore a learning-based approach to reconstruct HARDI from a smaller number of measurements in $q$-space. The approach aims to directly learn the mapping relationship between the measured and HARDI signals from the collecting HARDI acquisitions of other subjects.
Specifically, the mapping is represented as a 1D encoder-decoder convolutional neural network under the guidance of the compressed sensing (CS) theory for HARDI reconstruction. The proposed network architecture mainly consists of two parts: an encoder network produces the sparse coefficients and a decoder network yields a reconstruction result. Experiment results demonstrate we can robustly reconstruct HARDI signals with the accurate results and fast speed.
\end{abstract}
\section{Introduction}
\label{sec:intro}
High angular resolution diffusion imaging (HARDI) \cite{Qball} excels in detecting the orientational distribution of water diffusion in the cerebral tissue. However, it demands a higher amount of data measurements compared to diffusion tensor imaging (DTI). As the total scanning time increases linearly with the number of measurements \cite{Fastandaccurate}, HARDI-based analysis is currently deemed to be "too slow" to be used in clinical applications involving children or patients with dementia.

Currently, the high time-cost deficiency of HARDI can be overcome using the theory of compressed sensing (CS), which provides a framework to recover HARDI signals from a smaller number of measurements in $q$-space \cite{Fastandaccurate,ProbabilisticODF,Spatiallyregularized,Resolution,SHIMRM,SHICCPR}. We call these data measurements low angular resolution (LAR) signals.
This CS-based framework involves several steps in its pipeline. First, the relationship between the measurements and HARDI signal is modeled, and a LAR signal dictionary and a HARDI signal dictionary are reconstructed from multiple basis functions and corresponding diffusion-encoding gradient orientations. Then, the reduced measurements are encoded as sparse coefficients by the low resolution dictionary. Finally, the sparse coefficients are linearly mapped into HARDI signal by the high resolution dictionary. The CS-based works have shown competitive results, but there are two main problems in this framework. First, the relationship between the reduced measurements and the HARDI signal is hypothesized rather than learned from data, without using any statistical information. The performance of these CS-based algorithms degrades rapidly when the desired magnification factor is large or the number of available measurements is small \cite{viasparse}.
Second, the sparse coefficients in these works are usually obtained by solving a least squares optimization with $L_1$-norm regularization, which is very time-consuming and may have many solutions.

In this work, we investigate the possibility to learn the $q$-space signals lost in LAR acquisitions from the collecting HARDI acquisitions of other subjects. We explore a learning-based approach for recovering HARDI signals, in this case we can benefit from the statistical properties of the collecting HARDI acquisitions. Specifically,
we design a 1D convolutional neural network with the guidance of CS reconstruction algorithm. The network mainly consists of two parts: an encoder network that produces the sparse coefficients and a decoder network that produces a reconstruction result. We name the proposed network 1D Encoder-Decoder Convolutional Neural Network (1d-ED CNN).
The architecture of our network is based on autoencoder network with three important
improvements: (1) We design four input channels incorporating the information of the HARDI signal and its spherical coordinate, to make the 1D network is meaningful for HARDI signals; (2) We apply a strategy of randomly changing order to the measurement signals and their corresponding gradient coordinates, to fully learn the relationship between diffusion signals from different gradient orientations at the same spatial position; (3) We learn the network by a supervised paradigm, which enables us to directly learn the mapping between the LAR and HARDI signals.

The proposed 1d-ED CNN has several appealing properties. First, the entire pipeline is fully obtained through learning, without explicitly modeling the signal space or designing the dictionary. Second, we can benefit from the statistical properties of collecting high angular resolution signals from other subjects as training data. Third, our method is faster than a series of CS-based methods even on a CPU, because it is fully feed-forward and does not need to solve any optimization problem in usage.
Fourth, the property that the 4D real HARDI dataset usually contains very large number of voxels becomes an advantage in training a deep learning network, which is a burden for existing CS-based HARDI reconstruction methods.
We demonstrate the efficacy of our method on different numbers of reduced measurements compared with the existing CS-based reconstruction methods.

\section{Methods}
\subsection{CS Algorithm for HARDI Reconstruction}
\label{sec:format}
The relationship between diffusion signals from different diffusion gradient orientations in the spherical coordinate $\mathbf{q} \in  \mathbb{S}^2 :=\{\mathbf{v}\in\mathbb{R}^3|\|\mathbf{v}\|_2=1\}$ can be efficiently linearly represented using a dictionary and a vector of representation coefficients. For a fixed spatial position $\mathbf{r}_m \in \mathbb{R}^3$ , the HARDI signal vector $\mathbf{s}_m = [s(\mathbf{q_1};\mathbf{r}_m),s(\mathbf{q_2};\mathbf{r}_m),...,s(\mathbf{q_{K}};\mathbf{r}_m)]^T$ corresponding to $K_H$ diffusion-encoding orientations $\mathbf{Q}_H=\{\mathbf{q}_k\}_{k=1}^{K_H}$
can be represented by:
\begin{equation}\label{representation}
  \mathbf{s}_m=A_H(\mathbf{Q}_H)\mathbf{f}(\mathbf{r}_m)
\end{equation}
where $A_H(\mathbf{Q}_H)$ is HARDI signal dictionary composed by multiple basis functions of spherical ridgelets \cite{Fastandaccurate,Spatiallyregularized} or spherical wavelets \cite{ProbabilisticODF}.
$\mathbf{f}(\mathbf{r}_m)$ is a vector of representation coefficients which depends on the spatial coordinates $\mathbf{r}_m$. Consequently, one can recover the whole sphere by estimating the representation coefficients $\mathbf{f}_m$ (i.e., $\mathbf{f}(\mathbf{r}_m)$).
Given the diffusion signal vector $\mathbf{l}_m = [l(\mathbf{q_1};\mathbf{r}_m),l(\mathbf{q_2};\mathbf{r}_m),...,l(\mathbf{q_{K_L}};\\\mathbf{r}_m)]^T$ measured at a set of $K_L$ diffusion-encoding orientations $\mathbf{Q}_L=\{\mathbf{q}_k\}_{k=1}^{K_L}$ ($\mathbf{Q}_L$ is a subset of $\mathbf{Q}_H$), it can be modeled as:
\begin{equation}\label{models}
 \mathbf{l}_m = A_L(\mathbf{Q}_L)\mathbf{f}_m+\mathbf{e}_m
\end{equation}
where $A_L(\mathbf{Q}_L)$ is a measurement dictionary also composed by multiple basis functions and $e_k$ is the vector of corresponding measurement noise.

Due to the overcomplete dictionary and measured noise, then a practical
implication of this fact is that the coefficient vector $\mathbf{f}(\mathbf{r}_m)$ in Eq.(\ref{models}) is not unique. So this is an underdetermined problem.
Usually taking $L_1$-norm as the sparsity constraint is enforced on the coefficient vector. So we can recover $\mathbf{f}(\mathbf{r}_m)$ by solving the following optimization problem:
\begin{equation}\label{CS}
 \hat{\mathbf{f}}_m=\arg\min\{\|A_L(\mathbf{Q}_L)\mathbf{f}_m-\mathbf{l}_m\|_{L2}+\lambda\|\mathbf{f}_m\|_{L1}\}
\end{equation}

\begin{figure}
  \centering
  \includegraphics[width=0.5\textwidth]{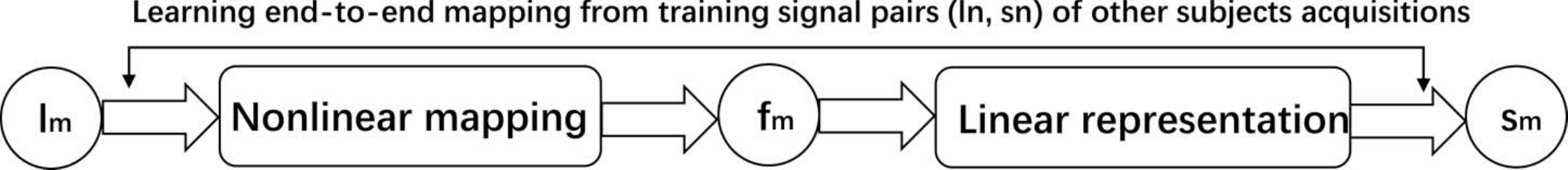}\\
  \caption{A learning-based approach that we want to directly learn mapping from the existing HARDI signals with the guidance of CS reconstruction algorithm.}\label{framework}
\end{figure}

So the mapping between $\mathbf{l}_m$ and $\mathbf{s}_m$ can be viewed as: first nonlinearly mapping the measurements $\mathbf{l}_m$
into a intact space to obtain the sparse representation coefficients $\mathbf{f}_m$ in Eq.(\ref{CS}) and then linearly mapping it into the high angular resolution representation space $\mathbf{s}_m$ in Eq.(\ref{representation}). Motivated by above pipeline, we investigate the possibility in Fig. \ref{framework} to directly learn an end-to-end mapping between the LAR signal $\mathbf{l}_m$ and original HARDI signal $\mathbf{s}_m$ from a set of signal pairs of other subjects.

\subsection{Network Architecture}
\begin{figure}
  \centering
  \includegraphics[width=0.5\textwidth]{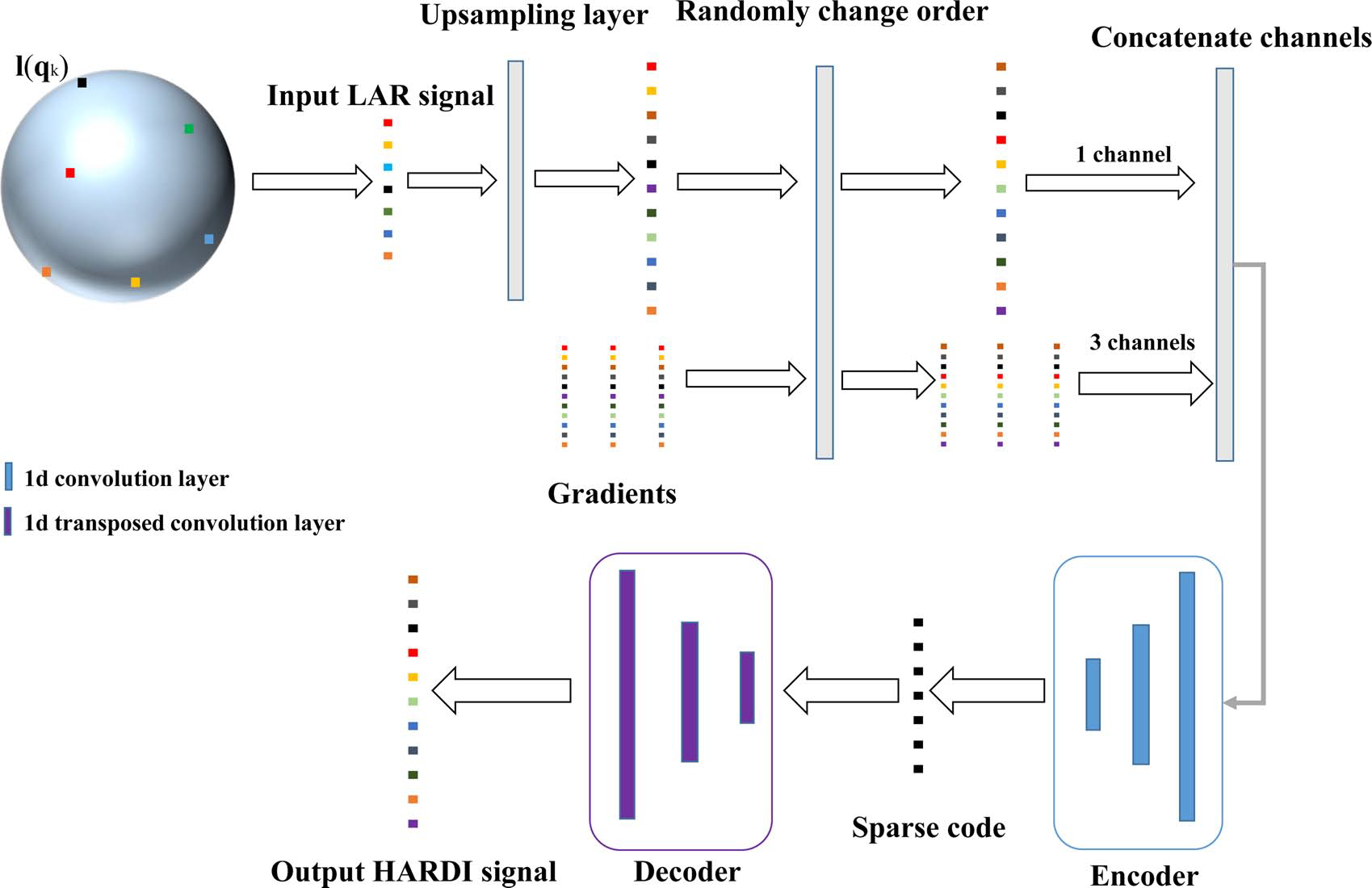}\\
  \caption{The architecture of the proposed network. Different colors in sequence represents the different gradient orientations in $q$-space.}\label{CNN}
\end{figure}

In this section, we design a 1D encoder-decoder convolutional neural network depicted in Fig. \ref{CNN} to learn the entire mapping 
in Fig. \ref{framework}.

\subsubsection{Input Layers}
For a measurement $\mathbf{l}_n\in \mathbb{R}^{K_L}$, we first upscale it to the same size of its ground truth HARDI $\mathbf{s}_n\in \mathbb{R}^{K_H}$ through upsampling layer or interpolation operation and its corresponding gradient orientation is $\mathbf{Q}_H =(\mathbf{Q_x},\mathbf{Q_y},\mathbf{Q_z})$. $\mathbf{Q}_x$, $\mathbf{Q}_y$ and $\mathbf{Q}_z$ are the point vectors in $x$, $y$ and $z$ axis with $\mathbf{Q_x},\mathbf{Q_y},\mathbf{Q_z}\in\mathbb{R}^{K_H}$.
Though we use a vector to represent the HARDI signal, we should note the diffusion signals $s(\mathbf{q_1};\mathbf{r}_m),s(\mathbf{q_2};\mathbf{r}_m),...,s(\mathbf{q_K};\mathbf{r}_m)$
or $l(\mathbf{q_1};\mathbf{r}_m),l(\mathbf{q_2};\mathbf{r}_m),\\...,l(\mathbf{q_{K_L}})$
is distributed on the surface of a sphere in $q$-space.
If we only input the signal vector without its $q$-space coordinate information, the spherical distribution of these diffusion signals is completely changed into a linear distribution. In this case, the 1D convolutional neural network is meaningless for the HARDI signal.
So we concatenate the $\mathbf{l}_n$, $\mathbf{Q}_x$, $\mathbf{Q}_y$ and $\mathbf{Q}_z$ as four channels of initial input $x_0$.
On the other hand, we should also note the fixed sequence order of the signal vector will prevent the 1D network to fully capture the relationship between diffusion signals from different gradient orientations. So we randomly disrupt the order of the diffusion signals in vector $\mathbf{l}_n$ and its corresponding spherical coordinate $\mathbf{Q}_x$, $\mathbf{Q}_y$ and $\mathbf{Q}_z$ as input $x_0$.

\subsubsection{1D Convolutional Encoder Network}
In the encoder stage, there are $I$ hidden layers to perform the nonlinear mapping. For $i$th hidden layer, it takes the input $x_{i-1}$ and maps it to $x_i$:
\begin{equation}\label{mapencoder}
x_i=\sigma(W^{i}_E\ast x_{i}+B_{i})
\end{equation}
where $W^{i}_E$, $B_{i}$ represent the filters and biases in encoder network, respectively, $\ast$ represents convolution operation and $\sigma$ is an element-wise activation function which allows us to transform the signal non-linearly. Because we expect to learn sparse representation, and we specify $\sigma$ as rectified linear unit.
After $I$ hidden layers transforming non-linearly, we can get the sparse code $\mathbf{f}_n$. The operation can be generalized as:
\begin{equation}\label{encoder}
\mathbf{f}_n=E(x_0;W^{1}_E,B_{1},...,W^{I}_E,B_{I})
\end{equation}

\subsubsection{1D Convolutional Decoder Network}
In the decoder stage, we also use $I$ symmetrical hidden layers to perform the linear mapping. For $i$th hidden layer, it takes the input $y_{i-1}$ and maps it to $y_i$:
\begin{equation}\label{mapencoder}
y_i= W^{i}_D\otimes y_{i-1}
\end{equation}
where $W^{i}_D$ represents the filters in decoder network and $\otimes$ represents deconvolution operation. In CS-based algorithm, the
measurement and HARDI signals dictionaries are composed by the same basis functions, so the size of $W^{i}_D$ is set the same to $W^{I+1-i}_E$
The initial input $y_{0}$ is sparse coefficients $\mathbf{f}_n$.
After $I$ hidden layers transforming linearly, we can get the reconstruction $\hat{\mathbf{s}}_n$. The operation can be generalized as:
\begin{equation}\label{decoder}
\hat{\mathbf{s}}_n=D(\mathbf{f}_n;W^{1}_D,...,W^{I}_D)
\end{equation}

\subsubsection{Loss Function}
Learning an end-to-end mapping function requires the estimation of parameters $\Theta = \{W^{1}_E,B_{1},...,W^{I}_E,B_{I},W^{1}_D,...,W^{I}_D\}$. This is achieved through minimizing the loss between the reconstructed signals and the corresponding ground truth HARDI signals through a supervised paradigm.
Given a training set including $N$ diffusion measurement sequences $\{\mathbf{l}_1,...,\mathbf{l}_N\}$ and the associated HARDI sequences $\{\mathbf{s}_1,...,\mathbf{s}_N\}$, we can get the corresponding reconstructed signals $\{\hat{\mathbf{s}}_1,...,\hat{\mathbf{s}}_N\}$ through the 1d-ED CNN. We use Normalized Mean Squared Error (NMSE) as the loss function:

\begin{equation}\label{MSE}
L(\Theta)=\frac{1}{N}\sum_{n=1}^{N}\frac{\|\hat{\mathbf{s}}_n-\mathbf{s}_n\|^2}{\|\mathbf{s}_n\|^2}
\end{equation}

\section{Experiments and Results}
\subsection{Datasets}
The real dMRI images of normal brains were from the HCP
\cite{HCP}, which have a spatial resolution of $1.25~mm \times 1.25~mm \times 1.25~mm$ with 90 diffusion weighting directions at $b= 2000 s/mm^2$. We use the diffusion signals normalized by the corresponding $\mathbf{B}_0$ image. The value of gradient directions also belong to [0,1]. Because the diffusion-encoded images were often contaminated by different levels of Rician noise, we used the LPCA filter in \cite{LPCA} to remove noise and used the preprocessed HARDI signals as the gold standard reference for quantitatively evaluating the results of the proposed method. We randomly selected 8000 diffusion signals as training set from 8 subjects and 2000 diffusion signals from other 2 subjects as testing data. For every HARDI signal $\mathbf{s}_n$ with the size of $90\times1$, we reduced it into $K_L\times1$ as measurement signal $\mathbf{l}_n$.

\subsection{Implementation Details}
In this experiment, we used three different $K_L$ values with $K_L = 30, 23, 18$ (the 1/3, 1/4, 1/5 of 90) in a range of values typical for DTI to validate the effectiveness of proposed method.
After the measurement signal was upsampled into the size of $90\times1$, we concatenated the $\mathbf{l}_n$ and the corresponding gradient directions
$\mathbf{Q}_x$, $\mathbf{Q}_y$ and $\mathbf{Q}_z$ as four channels of initial input $x_0$ with the strategy of randomly changing order.
In the encoder and decoder networks, we both used 3 hidden layers. All the kernel sizes of filters were set as $1\times9$. For 3 hidden layers in encoder network, the corresponding output channels were set to 400, 200 and 100, the stride sizes were set to 3, 3 and 2. In decoder network, the parameters were set symmetrically.
The learning rate was $0.001$ and batch size was 500.
For different values of $K_L$, we trained a specific network.
We compared our method with the CS-based algorithm RGD-CS in \cite{Spatiallyregularized} in terms of accuracy and speed. We also considered here the $L_2$-norm constraint in comparison because the $L_2$-solutions appear to be quite informative \cite{Fastandaccurate}. We named the $L_2$-norm constraint method RGD-$L_2$. All the methods were in python implementation and executed on a 2.00GHz Intel(R) Xeon(R) CPU.

\subsection{Quantitative Evaluation}

\begin{figure}[]
\centering
\subfloat[Original]{
\includegraphics[width=0.1\textwidth]{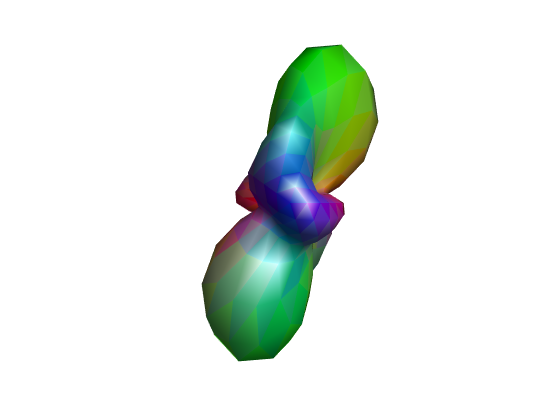}
}

\subfloat[RGD-$L_2$]{
\includegraphics[width=0.1\textwidth]{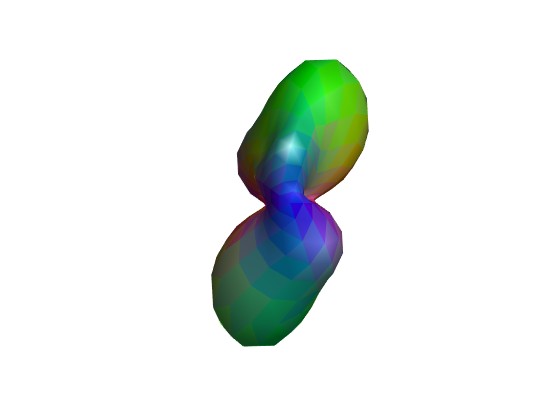}
}
\subfloat[RGD-CS]{
\includegraphics[width=0.1\textwidth]{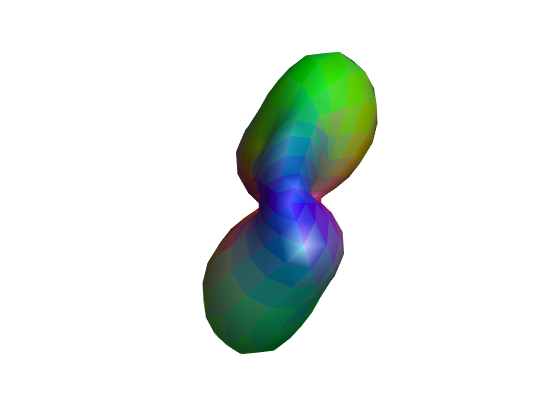}
}
\subfloat[Ours]{
\includegraphics[width=0.1\textwidth]{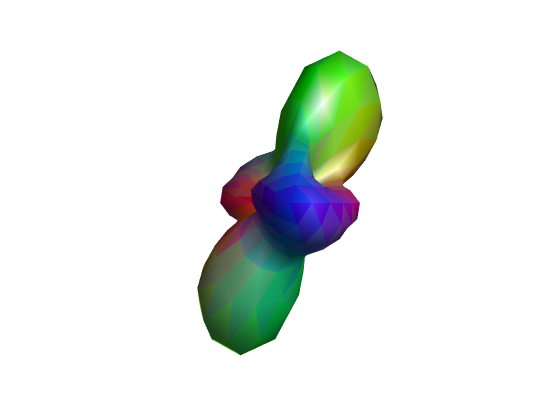}
}

\subfloat[RGD-$L_2$]{
\includegraphics[width=0.1\textwidth]{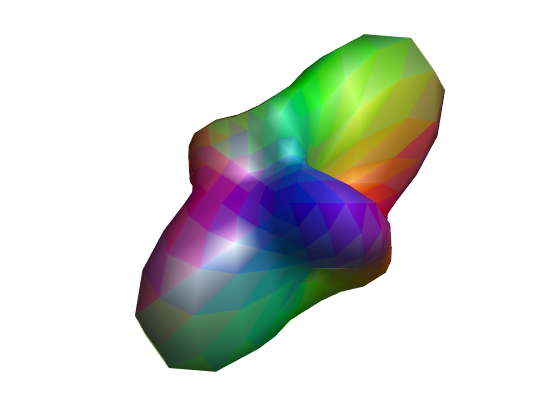}
}
\subfloat[RGD-CS]{
\includegraphics[width=0.1\textwidth]{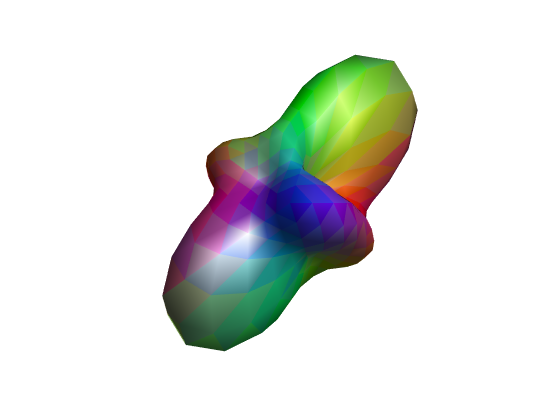}
}
\subfloat[Ours]{
\includegraphics[width=0.1\textwidth]{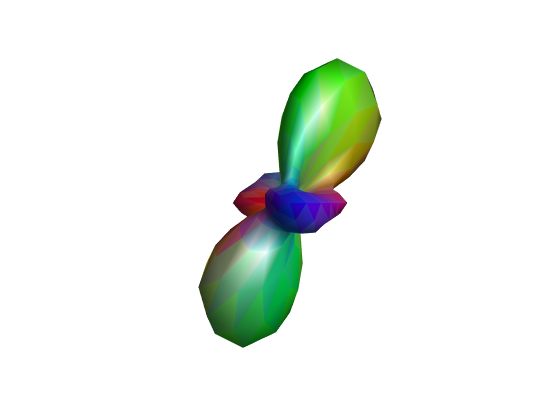}
}

\subfloat[RGD-$L_2$]{
\includegraphics[width=0.1\textwidth]{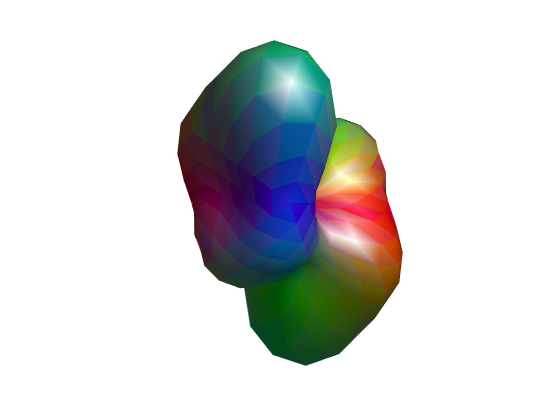}
}
\subfloat[RGD-CS]{
\includegraphics[width=0.1\textwidth]{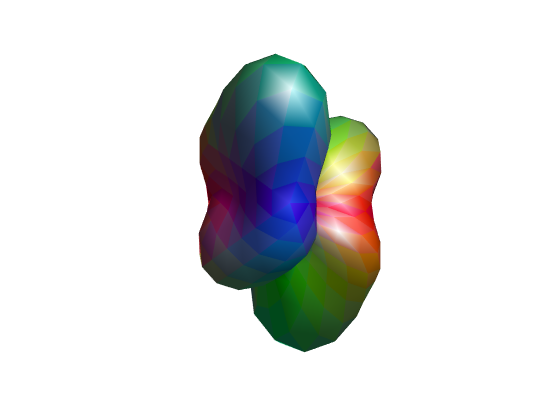}
}
\subfloat[Ours]{
\includegraphics[width=0.1\textwidth]{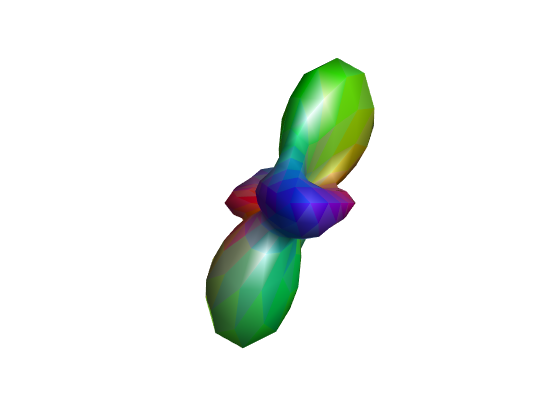}
}
\caption{\small The reconstructed ODF of different methods for different values of $K_L$. (b)-(d): $K_L = 30$; (e)-(g): $K_L = 23$; (h)-(j): $K_L = 18$.}
\label{b2000kspace2}
\end{figure}

As shown in Table \ref{NMSEresult2}, the proposed method yields the highest NMSE with all $K_L$ values. We also should note that the accuracy of RGD-CS and RGD-$L_2$ methods decreases rapidly as the $K_L$ value reduces, especially for the RGD-$L_2$ method. In contrast, the proposed method still produces promising results. This also demonstrats the robustness of our method. These observations are further visually depicted by orientation distribution function (ODF) images in Fig. \ref{b2000kspace2}. The ODF images were computed and visualized by the matlab codes in DSI studio \url{http://dsi-studio.labsolver.org/}. As can be observed, the ODF obtained by the proposed method are more close to the groundtruth.

\begin{table}
  \centering
   \caption{\small NMSE of different methods for different values of $K_L$.}
  \scalebox{0.55}{
  \begin{tabular}{|c|c|c|c|c|c|c|c|c|c|}
  \hline
  \multirow{2}{*}{Method}&\multicolumn{3}{c|}{$K_L=30$} & \multicolumn{3}{c|}{$K_L=23$}&\multicolumn{3}{c|}{$K_L=18$}\\
  \cline{2-10}
  & Min &Max &Average &Min &Max &Average &Min&Max&Average\\
  \hline
  RGD-$L_2$ & 0.0028 & 0.1872& 0.0187 &0.0096 &0.3856 & 0.0685&0.0408 &0.3153 & 0.1128 \\
  \hline
  RGD-CS & 0.0017 & 0.2207 & 0.0185 &0.0065 &0.3944 & 0.0404& 0.0169 &0.5652 & 0.0612\\
  \hline
  1D-SED CNN&\textbf{0.0017}& \textbf{0.1555} &\textbf{0.0154} &\textbf{0.0023}& \textbf{0.2117} &\textbf{0.0196}&\textbf{0.0021}& \textbf{0.1894} &\textbf{0.0199}\\
  \hline
\end{tabular}}
 \label{NMSEresult2}
\end{table}

\section{Conclusion}
In this paper, we investigated the possibility to learn the mapping relationship between the LAR and HARDI signals from the collecting HARDI acquisitions of other subjects. Specifically, we designed a novel 1d-ED CNN for HARDI signals reconstruction with the guidance of CS reconstruction algorithm. Experiment results demonstrate we can robustly reconstruct HARDI signals with the accurate results and fast speed.



\bibliographystyle{IEEEbib}
\bibliography{HARDI_ISBI}

\begin{thebibliography}{10}

\bibitem{Qball}
David~S. Tuch,
\newblock ``Q-ball imaging,''
\newblock {\em Magnetic Resonance in Medicine}, vol. 52, no. 6, pp. 1358--1372,
  2004.

\bibitem{Fastandaccurate}
Oleg Michailovich and Yogesh Rathi,
\newblock ``{Fast and accurate reconstruction of HARDI data using compressed
  sensing},''
\newblock in {\em International Conference on Medical Image Computing and
  Computer-Assisted Intervention}, 2010, pp. 607--614.

\bibitem{ProbabilisticODF}
Antonio Tristan-Vega and Carl-Fredrik Westin,
\newblock ``Probabilistic {ODF} estimation from reduced {HARDI} data with
  sparse regularization,''
\newblock in {\em International Conference on Medical Image Computing and
  Computer-Assisted Intervention}, 2011, pp. 182--190.

\bibitem{Spatiallyregularized}
Oleg Michailovich, Yogesh Rathi, and Sudipto Dolui,
\newblock ``Spatially regularized compressed sensing for high angular
  resolution diffusion imaging,''
\newblock {\em IEEE Transactions on Medical Imaging}, vol. 30, no. 5, pp.
  1100--1115, 2011.

\bibitem{Resolution}
Bennett~A. Landman, John~A. Bogovic, Hanlin Wan, Fatma El~Zahraa ElShahaby,
  Pierre-Louis Bazin, and Jerry~L. Prince,
\newblock ``Resolution of crossing fibers with constrained compressed sensing
  using diffusion tensor {MRI},''
\newblock {\em NeuroImage}, vol. 59, pp. 2175--2186, 2012.

\bibitem{SHIMRM}
Shi Yin, Xinge You, Xin Yang, Qinmu Peng, Ziqi Zhu, and Xiao-Yuan Jing,
\newblock ``A joint space-angle regularization approach for single 4d diffusion
  image super-resolution,''
\newblock {\em Magnetic Resonance in Medicine}, vol. 80, no. 5, pp. 2173--2187.

\bibitem{SHICCPR}
Shi Yin, Xinge You, Weiyong Xue, Bo~Li, Yue Zhao, Xiao-Yuan Jing, Patrick S.~P.
  Wang, and Yuanyan Tang,
\newblock ``A unified approach for spatial and angular super-resolution of
  diffusion tensor {MRI},''
\newblock in {\em Pattern Recognition: 7th Chinese Conference, CCPR 2016,
  Chengdu, China, November 5-7, 2016, Proceedings, Part II}, 2016, pp.
  312--324.

\bibitem{viasparse}
Jianchao Yang, John Wright, Thomas~S. Huang, and Yi~Ma,
\newblock ``Image super-resolution via sparse representation,''
\newblock {\em IEEE Transactions on Image Processing}, vol. 19, no. 11, pp.
  2861--2873, 2010.

\bibitem{HCP}
David C.~Van Essen, Stephen~M. Smith, Deanna~M. Barch, Timothy~E.J. Behrens,
  Essa Yacoub, Kamil Ugurbil, and {WU-Minn HCP Consortium},
\newblock ``The {WU-Minn} human connectome project: an overview,''
\newblock {\em NeuroImage}, vol. 80, pp. 62--79, 2013.

\bibitem{LPCA}
Jose~V Manjon, Pierrick Coupe, Luis Concha, Antonio Buades, D.~Louis Collins,
  and Montserrat Robles,
\newblock ``Diffusion weighted image denoising using overcomplete local
  {PCA},''
\newblock {\em PloS one}, vol. 8, no. 9, pp. e73021, 2013.

\end{thebibliography}
\end{document}